\documentclass{bmvc2k}

\title{\hspace{2ex}An ETF view of Dropout regularization}

\addauthor{\phantom{xxx}Dor Bank}{dorbank@gmail.com}{1}
\addauthor{\phantom{xxx}Raja Giryes}{raja@tauex.tau.ac.il}{1}

\addinstitution{
 School of Electrical Engineering \\
 The Iby and Aladar Fleischman Faculty of Engineering \\
 Tel Aviv University \\
 Tel Aviv, Israel
}

\runninghead{Bank and Giryes}{An ETF view of Dropout regularization}

\usepackage{graphics}
\usepackage{graphicx}
\usepackage{epstopdf}
\usepackage{lineno,hyperref}
\modulolinenumbers[5]
\usepackage{soul}
\usepackage{color}
\usepackage{tikz}
\usepackage{tikz-3dplot}
\usepackage{pgfplots}
\usepackage{pdfpages}
\usepackage{amsmath}
\usepackage{mathptmx}
\usepackage{gensymb}
\usepackage{amsfonts}
\usepackage{calrsfs}
\usepackage{array}
\usetikzlibrary{intersections}
\newcolumntype{P}[1]{>{\centering\arraybackslash}p{#1}}
\newcommand{\norm}[1]{\left\Vert#1\right\Vert}
\newcommand{\abs}[1]{\left\vert#1\right\vert}

\DeclareMathOperator{\argmin}{arg\,min}
\DeclareMathAlphabet{\pazocal}{OMS}{zplm}{m}{n}

\newcommand{\LL}{\pazocal{L}}
\newcommand{\D}{\pazocal{D}}
\newcommand{\diag}{\mathop{\mathrm{diag}}}

\begin{document}

\maketitle

\begin{abstract}
Dropout is a popular regularization technique in deep learning. Yet, the reason for its success is still not fully understood.
This paper provides a new interpretation of Dropout from a frame theory perspective. By drawing a connection to recent developments in analog channel coding, we suggest that for a certain family of autoencoders with a linear encoder, optimizing the encoder with dropout regularization leads to an equiangular tight frame (ETF). Since this optimization is non-convex, we add another regularization that promotes such structures by minimizing the cross-correlation between filters in the network. We demonstrate its applicability in convolutional and fully connected layers in both feed-forward and recurrent networks. All these results suggest that there is indeed a relationship between dropout and ETF structure of the regularized linear operations. 
\end{abstract}

\section{Introduction}

\hspace{2ex}Deep neural networks are powerful computational models that have been used extensively \phantom{xx}for solving problems in computer vision, speech recognition, natural language processing, \phantom{xx}and many other areas \cite{Krizhevsky2012ImageNetNetworks,Hinton2012DeepRecognition,Kim2014ConvolutionalClassification,Zhang16Towards,Voulodimos2018DeepReview}. The parameters of these architectures are \phantom{xx}learned from a given training set. Thus, regularization techniques for preventing overfitti-\\\phantom{xx}ng of the data are very much required \cite{Goodfellow16DeepLearningBook, Kukacka17Regularization}. Such methods include Batch Normalizati-\\\phantom{xx}on \cite{BatchNorm}, Weight decay \cite{WeightDecay}, $\ell_1$ regularization on the weights \cite{L1,L1_improper} and Jacobian regular-\\\phantom{xx}ization  \cite{Rifai11Contractive, Sokolic17Margin}.

One of the most popular strategies is $Dropout$, which randomly drops hidden nodes along with their connections at training time \cite{b5,Dropout}. During training, in each batch, nodes are kept with a probability $p$, which causes them to be eliminated with probability $q = 1 - p$ (with their corresponding input and output weights). The weights of the remaining nodes are trained by back-propagation regularly. At inference time, the outputs of the layer(s) on which Dropout was applied are multiplied by $p$. 
Though very useful, Dropouts explicit regularization is not fully understood yet. Such an understanding is required to exploit the full potential of Dropout, and to deepen our knowledge in neural networks.

\begin{figure}[t]
	\centering
\begingroup%
  \makeatletter%
  \providecommand\color[2][]{%
    \errmessage{(Inkscape) Color is used for the text in Inkscape, but the package 'color.sty' is not loaded}%
    \renewcommand\color[2][]{}%
  }%
  \providecommand\transparent[1]{%
    \errmessage{(Inkscape) Transparency is used (non-zero) for the text in Inkscape, but the package 'transparent.sty' is not loaded}%
    \renewcommand\transparent[1]{}%
  }%
  \providecommand\rotatebox[2]{#2}%
  \ifx\svgwidth\undefined%
    \setlength{\unitlength}{148.40805626bp}%
    \ifx\svgscale\undefined%
      \relax%
    \else%
      \setlength{\unitlength}{\unitlength * \real{\svgscale}}%
    \fi%
  \else%
    \setlength{\unitlength}{\svgwidth}%
  \fi%
  \global\let\svgwidth\undefined%
  \global\let\svgscale\undefined%
  \makeatother%
  \begin{picture}(1,0.28247945)%
    \put(0,0){\includegraphics[width=\unitlength]{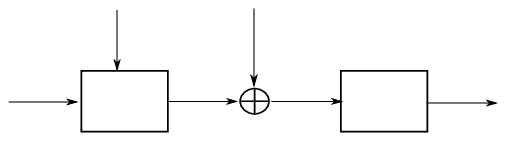}}%
    \put(0.11747132,0.31040551){\color[rgb]{0,0,0}\makebox(0,0)[lt]{\begin{minipage}{0.327928\unitlength}\raggedright $Dropout(p)$\end{minipage}}}%
    \put(0.48581537,0.2917323){\color[rgb]{0,0,0}\makebox(0,0)[lt]{\begin{minipage}{0.09767706\unitlength}\raggedright $w$\end{minipage}}}%
    \put(0.20974552,0.08445851){\color[rgb]{0,0,0}\makebox(0,0)[lt]{\begin{minipage}{0.08764425\unitlength}\raggedright $E$\end{minipage}}}%
    \put(10.88749057,-3.60983829){\color[rgb]{0,0,0}\makebox(0,0)[lt]{\begin{minipage}{0.6857735\unitlength}\raggedright \end{minipage}}}%
    \put(1.01182019,0.0813754){\color[rgb]{0,0,0}\makebox(0,0)[lt]{\begin{minipage}{0.27373592\unitlength}\raggedright $\hat {x}$\end{minipage}}}%
    \put(2.7275004,-1.05676075){\color[rgb]{0,0,0}\makebox(0,0)[lb]{\smash{}}}%
    \put(-0.05608764,0.05466835){\color[rgb]{0,0,0}\makebox(0,0)[lb]{\smash{$x$}}}%
    \put(0.58997648,0.09060868){\color[rgb]{0,0,0}\makebox(0,0)[lb]{\smash{$y$}}}%
    \put(0.74286154,0.08004692){\color[rgb]{0,0,0}\makebox(0,0)[lt]{\begin{minipage}{0.28589345\unitlength}\raggedright $D$\end{minipage}}}%
  \end{picture}%
\endgroup%
\hspace{12mm}%
\begingroup%
  \makeatletter%
  \providecommand\color[2][]{%
    \errmessage{(Inkscape) Color is used for the text in Inkscape, but the package 'color.sty' is not loaded}%
    \renewcommand\color[2][]{}%
  }%
  \providecommand\transparent[1]{%
    \errmessage{(Inkscape) Transparency is used (non-zero) for the text in Inkscape, but the package 'transparent.sty' is not loaded}%
    \renewcommand\transparent[1]{}%
  }%
  \providecommand\rotatebox[2]{#2}%
  \ifx\svgwidth\undefined%
    \setlength{\unitlength}{148.40805626bp}%
    \ifx\svgscale\undefined%
      \relax%
    \else%
      \setlength{\unitlength}{\unitlength * \real{\svgscale}}%
    \fi%
  \else%
    \setlength{\unitlength}{\svgwidth}%
  \fi%
  \global\let\svgwidth\undefined%
  \global\let\svgscale\undefined%
  \makeatother%
  \begin{picture}(1,0.38247945)%
    \put(0,0){\includegraphics[width=\unitlength]{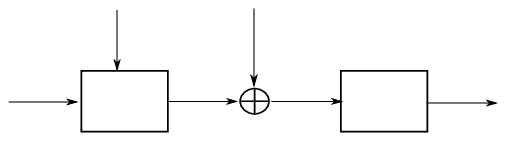}}%
    \put(0.17795273,0.31040551){\color[rgb]{0,0,0}\makebox(0,0)[lt]{\begin{minipage}{0.15782402\unitlength}\raggedright $S(p)$\end{minipage}}}%
    \put(0.48581537,0.2917323){\color[rgb]{0,0,0}\makebox(0,0)[lt]{\begin{minipage}{0.09767706\unitlength}\raggedright $w$\end{minipage}}}%
    \put(0.20974552,0.08445851){\color[rgb]{0,0,0}\makebox(0,0)[lt]{\begin{minipage}{0.08764425\unitlength}\raggedright $F$\end{minipage}}}%
    \put(10.88749057,-3.60983829){\color[rgb]{0,0,0}\makebox(0,0)[lt]{\begin{minipage}{0.6857735\unitlength}\raggedright \end{minipage}}}%
    \put(1.01182019,0.0813754){\color[rgb]{0,0,0}\makebox(0,0)[lt]{\begin{minipage}{0.27373592\unitlength}\raggedright $\hat {x}$\end{minipage}}}%
    \put(2.7275004,-1.05676075){\color[rgb]{0,0,0}\makebox(0,0)[lb]{\smash{}}}%
    \put(-0.05608764,0.05466835){\color[rgb]{0,0,0}\makebox(0,0)[lb]{\smash{$x$}}}%
    \put(0.58997648,0.09060868){\color[rgb]{0,0,0}\makebox(0,0)[lb]{\smash{$y$}}}%
    \put(0.73530135,0.0876071){\color[rgb]{0,0,0}\makebox(0,0)[lt]{\begin{minipage}{0.28589345\unitlength}\raggedright $F^\dag_s$\end{minipage}}}%
  \end{picture}%
\endgroup%

	\caption{(left) a DAE variant (see \cite{denoising_AE}) with a linear encoder; (right) a signal encoding scheme in an analog channel with a decoder that performs least squares based inversion. $S(p)$ is a sampling pattern with a sampling ratio $p$ and $w \in \mathbb{R}^{p \cdot n}$ is an additive noise.
	}  \label{fig:process}. 
\end{figure}

This work approaches Dropout from a signal processing and information theory perspective. It draws a connection between Dropout in a denoising autoencoder (DAE) and signal recovery from erasures in the analog domain (see Fig.~\ref{fig:process}). In this ``analog coding'' problem, a signal passes through an encoder $A$ and then disrupted by an additive noise and part of its values are nullified. Once received, it is recovered by passing through a decoder $B$.  
The goal is to find the pair $(A,B)$, which recovers the input signal with a minimal $\ell_2$ error.

To draw a connection to Dropout, we make the following steps.
First, we examine a specific case, where the encoding $A$ is performed by a (linear) matrix multiplication $F$, and the recovery is done by solving a least squares problem with the given measurements and $F_{s}$, the subset of columns from the matrix $F$ corresponding to the kept measurements.
It has been suggested in a recent work that frames with MANOVA distribution \cite{MANOVA}, minimize the expected $\ell_2$ error in this setting \cite{Marina_AnalogCoding}. Though not proven formally, various empirical measurements lead to the conjecture that ETFs have  MANOVA distribution in their sub-matrices, and thus minimize the $\ell_2$ error in the above setup \cite{Marina_ETFManova,Marina_FrameMoments}.

Next, we draw a relationship to DAE (briefly illustrated in Fig.~\ref{fig:process}). 
Considering an autoencoder with a linear  encoder and a Dropout regularization applied on it, we get a very similar structure to the analog coding problem. Thus, if the decoder solves the least squares problem, then an ETF is likely to be a global minimum in the encoder optimization. 

Last, we notice that the representation learned by autoencoders may be used for classification, e.g., in a semi-supervised learning setup, where the learned encoder serves as a feature extractor. This leads to the conjecture that promoting structure of an ETF in some layers of the network might turn useful for classification tasks as well.  This provides a first step towards using frame theory for understanding and improving neural networks.

We support our claim by experiments done on various data-sets for image classification and word level prediction. We measure the effect of the ETF regularization when used as a sole regularizer, and when combined with Dropout. For fully connected (FC) layers, we promote an ETF structure for the weight matrix directly by reducing the correlation between its rows. We demonstrate this regularization for both feed-forward and recurrent (LSTM) networks. For convolutional layers, we do not use their corresponding Toeplitz matrix. Instead, for simplicity, the coherence between the convolution kernels is minimized.


\section{Related works}

This section discusses some previous works that analyze dropout. A detailed description of dropout and autoencoders appears in the sup. mat. 

One disadvantage of Dropout, is that it slows down the training time. Wang and Manning, have implied that Dropout makes a Monte Carlo assessment of the layers output and thus reduced the training time \cite{fast_dropout}.
Frazier-Logue and Hanson  suggest that Dropout is just a special case of a stochastic delta rule, where each weight is parameterized as a random variable with a mean and variance of its own. 
Their method leads to faster convergence than using Dropout \cite{delta_rule}.
Hara et al. compared training with Dropout to ensemble learning, where several sub-networks are learned independently, and then the final result is an aggregation of all of them \cite{Dropout_ensemble}. 
Baldi et al. \cite{b4, b4a} introduced a general formalism for studying Dropout in networks with the sigmoid activation function. They showed that for a shallow network the expected output of a network with dropout can be approximated via the weighted geometric mean of the network outputs.
Wager et al. analyzed Dropout applied to the logistic loss for generalized linear models (GLM) \cite{NIPS2013_4882}. 
They claim that Dropout is similar to applying $\ell_2$ regularization, where each squared weight is normalized using the Fisher information matrix. 

Helmond and Lond derived a sufficient condition to guarantee a unique minimizer for a loss function that uses Dropout \cite{b12}. To differentiate between the bias induced by Dropout and $\ell_2$ regularization, they provide examples for input data distributions for which the error achieved by Dropout is lower than the one of $\ell_2$, and examples for the opposite case.
Wager et al. showed for a generative Poisson topic model with long documents that Dropout training improves the exponent in the generalization bound for empirical risk minimization \cite{NIPS2014_5502}.
Cavazza et al. discussed the equivalence between Dropout and a fully deterministic model for Matrix Factorization in which the factors are regularized by the sum of products of the squared Euclidean norms of the columns of the matrix \cite{b11}. Pal et al. showed equivalence between Dropout and DropConnect \cite{DropConnect}, and that for single
hidden-layer linear networks, DropBlock \cite{DropBlock} induces spectral
k-support norm regularization, and promotes solutions that
are low-rank and have factors with equal norm \cite{structured_dropout}.
Tang et al. proposed DisOut, a method for feature distortion based on the network layers empirical Rademacher complexity \cite{DisOut}.

Gal and Ghahramani use Dropout to measure the uncertainty of a network. They approximate the likelihood functions with Monte Carlo sampling done via Dropout \cite{Dropout_Bayesian_approx}.

The two methods most related to our work are the one by Mianjy et al. \cite{b2} and DeCov \cite{DeCov} described in detail in the sup. mat. 
The first studies the implicit bias of Dropout \cite{b2}. It focuses on the case of a shallow autoencoder with a single hidden layer. It draws a relationship between the norms in the encoder and the decoder showing that they need to be equalized.   
The second by Cogswell et al. \cite{DeCov} uses the fact that Dropout leads to less correlated features and thus suggests to regularize the covariance of the features with respect to the training data.
Hereafter, we compare our theory to the one of Mianjy et al. \cite{b2} and show that our proposed regularization method enforces jointly equalized matrices when performed in a linear autoencoder, and mention the connection between our work and Decov.

\section{Signal reconstruction from a frame representation}
\label{setup}
We now address a notorious problem in information theory: Signal reconstruction from a frame representation with erasures, as illustrated in Fig.~\ref{fig:process}. Later on, we shall use its resemblance to autoencoders.
Consider the signal vector $x \in \mathbb{R}^m$ and a frame $F$.
First, the vector is encoded by $F$, i.e., yielding $xF$, which is then transmitted in an analog channel. In the channel, part of the values are nullified with probability $p$, and then the remaining values are disrupted by an additive white Gaussian noise (AWGN). 

Notice that nullifying the values in $xF$ with probability $p$ is equivalent to removing columns from $F$ with probability $p$ and then multiplying it with $x$.
Denote by $S(p)$ the pattern that defines which vectors of $F$ are used, with respect to the probability $p$, and by $F_s$ the sub-matrix of $F$ with the vectors corresponding to $S(p)$. Then the resulted vector after the addition of the AWGN $w$ is defined as
\begin{equation}
y = xF_s+w.
\end{equation}
In order to recover the input from $y$, one may use the least square solution
\begin{eqnarray}
\hat{x} = \argmin_{\tilde{x}}\norm{y - \tilde{x} F_s}_2^2 =  yF^{\dag}_s,
\end{eqnarray} 
where $F^{\dag}_s$ is the pseudo-inverse of $F_s$. 
Thus, if one wishes to optimize $F$ for minimizing the reconstruction error in the $\ell_2$ sense, the target objective is:
\begin{eqnarray}\label{eq:target}
\argmin_F \mathbb{E}\norm{x-\hat{x}}_2^2 =  \argmin_F \mathbb{E}\norm{ x-yF^{\dag}_s }_2^2 ,
\end{eqnarray}
where the expectation is with respect to the noise variable $w$, the distribution of the input variable $x$, and the sampling vector $S(p)$.

{\bf Frames for signal encoding.}
A number of works have studied the problem of reconstruction from erasures in the setup presented in Fig.~\ref{fig:process}  (see for example \cite{UTF, erase1, erase3, Larson2015}). As part of it, the usage of frames as encoders was vastly explored.
Frames, or overcomplete bases, are $m \times n$ matrices with rank $m$, where $n>m$.
They are widely used in various applications of communication, signal processing, and harmonic analysis \cite{frames1, frames2, frames3, frames4}. For example, they are often used for sampling techniques to analyze and digitize signals and images when they are represented as vectors or functions in a Hilbert space \cite{Eldar15Sampling}.

There is also a great interest in finding frames with favorable properties that hold for random subsets of their columns \cite{SubsetInterest}. 
One popular type of frames is tight frames. A frame $F$ of dimensions $m\times n$ is a tight frame iff $FF^T = c\cdot I_m$ for some constant c. In \cite{TF}, they have been shown to be useful for quantization. 

{\bf Equiangular tight frames.}
An interesting sub-group of tight frames are  ETFs.
The Gram matrix of a frame $F$ is defined by $G_F = F^T F$ and contains outside its diagonal the cross-correlation values between the columns of the frame $F$, i.e., $G_{i,j}$ contains the cross-correlation value between the $i$th and $j$th columns of $F$.
The Welch bound \cite{b6} provides a universal lower bound on the mean and maximal absolute value of the cross-correlations between the frame vectors.
A frame that achieves the Welch lower bound on the maximal absolute cross-correlation value is an ETF.  
The Gram matrix $G_{ETF}$ of a $m \times n$ ETF satisfies:
\begin{equation}\label{eq:gram_matrix}
\abs{(G_{ETF})_{i,j}} =  \begin{cases}
\begin{array}{c}
1\\
\frac{n-m}{(n-1)m}
\end{array} & \begin{array}{c}
i=j\\
else.
\end{array}\end{cases}
\end{equation}
 Intuitively, the $n$ vectors of an ETF are spread uniformly across an $m$ dimensional space with an angle $\theta = \arccos{\sqrt{\frac{n-m}{(n-1)m}}}$ between them.
 The maximal off-diagonal value in the Gram matrix is denoted the mutual coherence \cite{Donoho02Optimally} or simply the coherence value.


It was demonstrated that frames that reach the Welch bound (also known as Equiangular Tight Frames(ETF)), have MANOVA distribution \cite{Marina_ETFManova}. The eigenvalue distribution of the submatrices of an ETF is shown empirically to resemble the MANOVA distribution. We provide a brief intuition here and more details in the sup. mat.
Note that minimizing the estimation error at the decoder output is equivalent to minimizing $\mathbb{E}[Tr(F^T_sF_s)^{-1}]$ because
\begin{align}
\label{eq:MSE_equivalence}
    &\argmin_F \mathbb{E}\norm{x-\hat{x}}_2^2 =\argmin_F \mathbb{E}\norm{x-F^{\dag}_sFx+F^{\dag}_sw}_2^2 
    =\argmin_F \mathbb{E}\norm{F^{\dag}_sw}_2 
 \\ \nonumber
    &=\argmin_F \mathbb{E}(Tr(F^{\dag T}_sF^{\dag}_sww^T)) =\argmin_F \sigma^2_w \cdot \mathbb{E}(Tr(F^T_sF_s)^{-1}) 
    =\argmin_F \mathbb{E}(Tr(F^T_sF_s)^{-1}).
\end{align}

Assuming the frame columns are normalized and $F_s$ has $k$ columns, then $\mathbb{E}[Tr(F^T_sF_s)]$ is independent of $S$ and equals to $=\sum_{i=1}^k{\lambda_i}$, where $\lambda_k$ is the $k^{th}$ eigenvalue of $F$. Thus, the minimization objective in Eq.~\eqref{eq:MSE_equivalence} becomes $\mathbb{E}\left[\sum_{i=1}^k{\frac{1}{\lambda_i}}\right]$. In this case, it is clear that the best possible distribution is $p(\lambda)=\delta(\lambda)$, i.e., each sub-matrix is unitary. Yet, this is impossible to maintain for over-complete frames for which all the sub-matrices cannot be unitary.

To assess that Manova is the optimal choice, various popular random matrices with known distributions were tested \cite{Marina_ETFManova}. These include Low pass frames with Vandermonde distribution \cite{Vandermonde}, Gaussian frames that obeys  the Marchenko-Pastur distribution \cite{Marina_ETFManova}, and frames whose distribution resembles the MANOVA distribution such as ETF, Random Fourier, and Haar \cite{Marina_ETFManova}. MANOVA was shown to be the distribution closest to $\delta(\lambda)$.
Thus, overall we get the conjecture that ETFs are the global minimum for the settings of Eq.~\eqref{eq:target}.

\section{An ETF perspective of Dropout}\label{sec:perspective}

Having the problem of reconstruction of a signal with erasures stated, we turn to draw a relationship between it and optimizing a neural network with dropout. In particular, we focus mainly on the relationship to autoencoders. 

\subsection{The relationship between Dropout and ETF}\label{sec:relationship}
Notice the great resemblance between a denoising autoencoder (DAE) with linear encoder and Dropout applied on it and the analog coding problem, as illustrated in Fig.~\ref{fig:process}. Though in standard DAE the noise is added at the input, in the DAE we present here, we put the noise at the output of the encoder, as suggested in \cite{denoising_AE}. There is a similarity between the two models in the linear case as adding noise at the output of the decoder is equivalent to adding noise at its input with a covariance matrix equal to the pseduo-inverse of the decoder.

Given the above information, in the case that the encoder is linear and the decoder calculates the least squares solution, we conjecture that the global minimum of training with Gaussian distributed data and noise, and Dropout on the encoder should be an ETF for the encoder (or very close to it if the setting slightly changes). 

Notice that for a given Dropout/erasure pattern, the decoder is a linear operation. Since the encoder is also linear, the autoencoder with the fixed pattern becomes a shallow linear autoencoder as used in \cite{b2} (See sup. mat.). In that work it is claimed that Dropout induces the matrices of such a shallow linear autoencoder to be jointly equalized. In our case, the optimal encoder is claimed to be an ETF and thus the linear encoder and decoder in the linear autoencoder induced by the Dropout  are a sub-matrix of an ETF and its pseudo-inverse, respectively. Interestingly, it turns out that this pair is indeed jointly equalized, which corresponds with the theory derived in  \cite{b2}. Notice that this is not exactly the result derived in that work, since unlike their assumption that the decoder is linear, here it is non-linear (it is linear only given a specific erasure pattern). Thus, this relationship requires further study.

To examine the relationship between Dropout and ETFs, we set an experiment with an autoencoder that has a similar structure to the analog coding problem setup described in Section~\ref{setup}. The encoder $A$ in this network is a linear one, represented by a randomly initialized matrix. Specifically, we use a matrix $A$ of size $75 \times 150$. 

For the decoder, we do not use $A^\dag_s$ since it is hard to calculate its derivative with respect to $A$ during training. Instead, we use the fact that the pseudo inverse is the least squares solution and perform ten iterations of gradient descent $\hat{x}^{i+1} = \hat{x}^{i}-\mu A^T_s(A_s\hat{x}^{i}-y)$, where $\hat{x}=\hat{x}^{9}$ and $x = \hat{x}^{0}$. The learning rate $\mu$ is the inverse of the largest eigenvalue of the Gram matrix $A^T_sA_s$ as in \cite{EigenValue}. For the sample pattern $S(p)$, we simply apply Dropout on the encoder.

The input signals are generated as i.i.d. Gaussian vectors with a standard deviation of $1$ and the noise is generated with the same distribution but with a standard deviation of $0.001$.

\begin{figure}
    \includegraphics[scale = 0.4]{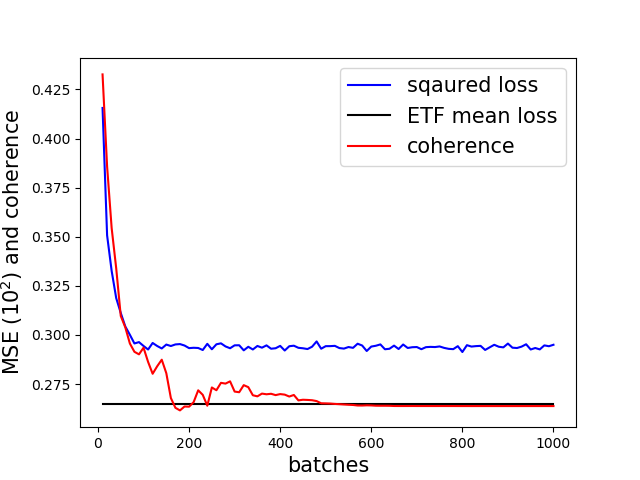}
    \includegraphics[scale = 0.4]{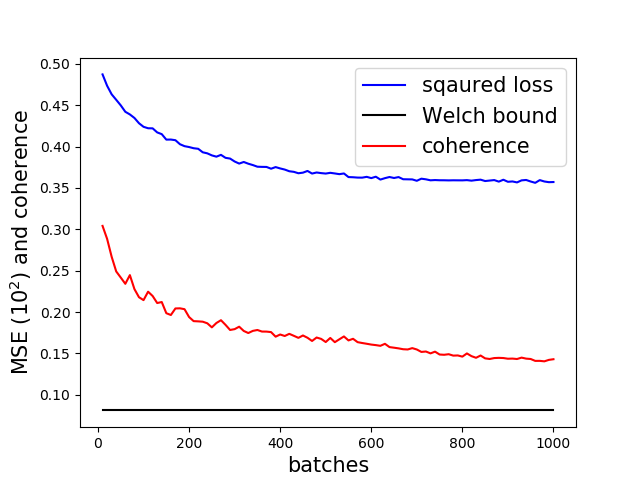}
    \caption{Training linear DAE with infinite data: plots of the coherence and the squared error as a function of batches. The error is scaled by 100 to fit with coherence in the same plot. Left: optimizing over the MSE. Right: optimizing over the Encoder coherence.}
    \label{fig:infinite_data}
\end{figure}

\begin{figure}
\centering
\begin{tikzpicture}
\begin{axis}
        [
        ,width=7cm
        ,height=5cm
        ,ymin = 0.0, ymax = 0.8
        ,xlabel=$\beta$
        ,ylabel= {$\ell_2$ error ($10^2$) and coherence}
        ,xtick=data,
        ,xticklabels={$0$,$10^{-4}$,$10^{-3}$,$10^{-2}$,$10^{-1}$, $10^{0}$, $10^{1}$},
        axis lines=left
        ,legend pos=outer north east]
        \addplot+[sharp plot] coordinates
        {(0,0.43) (1,0.43) (2,0.42) (3,0.37) (4,0.35) (5,0.34) (6,0.39)};
        \addlegendentry{minimal test MSE}
        \addplot+[sharp plot] coordinates
        {(0,0.47) (1,0.47) (2,0.46) (3,0.37) (4,0.35) (5,0.43) (6,0.78)};
        \addlegendentry{final test MSE}
        \addplot+[sharp plot] coordinates
        {(0,0.47) (1,0.3) (2,0.22) (3,0.16) (4,0.148) (5,0.146) (6,0.145)};
        \addlegendentry{Encoder coherence}        
        \addplot[dashed, black] coordinates {(0,0.27) (6,0.27)};
        \addlegendentry{ETF MSE}
        \addplot[dotted, black] coordinates {(0,0.1) (6,0.1)};
        \addlegendentry{ETF coherence}

\end{axis}
\end{tikzpicture}
\caption{Training linear DAE with finite data: Plots of the squared error of a DAE and the coherence of its encoder as a function of $\beta$. The error is scaled by 100 to put it with the coherence in the same plot. For each $\beta$, the minimal and final MSE is measured and compared to the one of an ETF with a pseudo inverse decoder.}
\label{fig:coherence_regularization}
\end{figure}
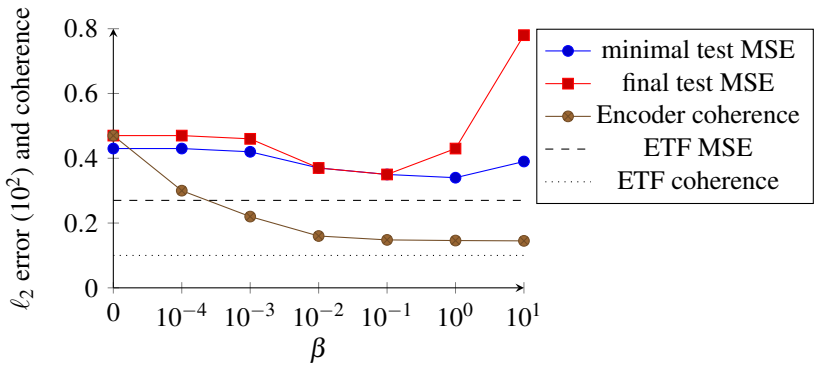

The experiment is performed in two different settings: the first includes an infinite amount of data, and the second deals with a finite and limited one.
In the infinite data case, we seek to find correspondence between the encoder coherence and the squared loss. We use an "online" learning setup with 100 signals per batch. First, we optimize over the squared loss and measure the coherence. Second, we optimize over the following "Coherence loss":
\begin{equation}\label{eq:Co_Loss}
CL = \|{A^TA}-\abs{G_{ETF}}\|_\infty,
\end{equation}
where $\abs{\cdot}$ is an element-wise absolute value, and $\norm{\cdot}_\infty$ returns the maximum absolute value in the matrix.
As can be seen in Fig.~\ref{fig:infinite_data}, the coherence and the reconstruction loss are closely related. Notice that coherence minimization induces a MSE reduction and vice versa. This validates our claim on the relationship between the two. Indeed, the coherence does not reach the Welch bound, and the error is much higher than the one of an ETF. We conjecture that this is mainly due to the  non-convexity of the problem. 

For limited data, regularization of the coherence is considered such that the new loss is
\begin{equation}\label{eq:regularization}
L = MSE + \beta \cdot  CL,
\end{equation}
where $\beta$ is the regularization coefficient. Note that this term encourages getting an ETF-like structure. We train the autoencoder with this new regularization over several values of $\beta$. We use a training set of 100 signals, where the training phase includes randomizing the noise vector and the sampling pattern in each batch. The test set size is chosen as 5000 to accurately measure the test error. We train the model for 300 epochs in which the minimal and final test errors are measured, along with the final coherence.

It is known that regularization techniques increase the bias of a model. If successful, they reduce the models variance such that the total error is reduced, and thus prevent overfitting. Therefore, we expect low regularization coefficients to have little effect on the performance, large ones to perform poorly due to high bias, and for a specific range to increase the training error while decreasing the test error.
Fig.~\ref{fig:coherence_regularization} shows that until a certain value, both the error and the coherence diminish as $\beta$ increases. High $\beta$ values (in this case - higher than $0.1$), result with optimization difficulties. This demonstrates that adding this term indeed helps in improving the convergence of the encoding frame closer to the desired ''global minimum''.

\subsection{Promoting an ETF structure in general neural networks}\label{sec:coherence}

Recalling the setting of Section~\ref{setup}, notice that the encoding part is exactly equivalent to a FC layer in a  neural network, where the frame $F$ plays the roll of the weight matrix, and the nullification with probability $p$ acts as Dropout. Though the specific setup discussed here is more relevant to autoencoders, we believe that the new understandings of Dropout may be carried also to more general neural networks. 
Inspired by the usage of autoencoders for classification, we conjecture that ETFs may be helpful for regularization also in other tasks beyond signal recovery, e.g. for classification as we demonstrate hereafter.

Since there are infinitely many ETFs, we do not want to regularize a layer towards a specific one. Moreover, we do not always have an ETF construction for every combination of $m$ and $n$. Yet, the structure of the Gram matrix is easily accessible and is the same for all ETFs that have the same value of $m$ and $n$.

For these reasons, and the ones specified in Section \ref{sec:relationship}, we adopt the "ETF similarity" term presented in Eq. \eqref{eq:regularization} also for general neural networks and in particular for ones performing classification tasks.
Notice that in the case where $m>n$, all vectors can be independent, and we penalize the distance from $I_m$, which is the same as reducing the magnitude of the off-diagonal entries of $A^T A$. In the case of convolution, we regularize the coherence between the convolution kernels (we justify this selection in the sup. mat.).

In an LSTM cell, we have four different FC gates: One to create a new state vector; one to create a $forget$ vector, which decides how much to keep from the old state; One for an $input$ vector,
which decides how much to keep from the new state; and one for the cell's $output$. We promote ETF-like matrices on each one of them separately, since we do not want to impose low coherence between the vectors of the different FC layers (We may still want that the same filter will be used in the different gates.).

Interestingly, our proposed coherence based regularization technique may also be motivated by the sparse coding theory, where it is well known that it is easier to recover the sparse representation of a vector from a matrix that has a low coherence \cite{Donoho02Optimally, Elad07Optimized}. In a recent work, it has been shown that the layers of a convolutional neural network may be viewed as stages for reconstructing the sparse representation of the input  \cite{Papyan18Theoretical}. Moreover, recovery guarantees have been developed based on the coherence showing that a smaller coherence leads to better reconstruction of the sparse representation of the input by the network \cite{Papyan17Convolutional}. While that work focuses mainly on convolutional layers, it definitely provides another motivation for our new regularization technique. This is especially true since in classical sparse coding the coherence is also used with regular matrices (equivalent to the weights in the FC layers).

Practically, there are few ways to promote a matrix $A$ to be an ETF-like, i.e., making its coherence as small as possible. 
We focus on three of them: minimizing the sum of squares of $\abs{A^TA}-\abs{G_{ETF}}$, the sum of absolute values and the maximal value, which is equivalent to minimizing the coherence of $A$ as in \eqref{eq:regularization}. Notice that minimizing the sum of absolute values is similar to the approach used in \cite{Elad07Optimized} for minimizing the coherence in a dictionary by reducing the average absolute value of the cross-correlations between its columns. Another approach proposed in \cite{Carvajalino09Learning} relies on a spectral decomposition of $A$. Though it is shown to be more effective than the one in \cite{Elad07Optimized}, it is too computationally demanding for using it with a neural network training and thus we focus only on techniques that minimize the coherence directly.
In addition, by assuming that the inputs are Gaussian, it is possible to minimize the cross-correlations of the columns, which partially coincides with the DeCov method \cite{DeCov} that penalizes the activation cross correlations. 

We compare the three regularization options above with a classification network for the Fashion MNIST dataset. We regularize the FC layer in a LeNet5 type network (the exact  settings are detailed in Section~\ref{Experiments}).  
Table \ref{table:FashionMnistLosses} presents the classification accuracy on the test set. We select for each regularization strategy its own optimal parameter $\beta$. This table suggests that minimizing the coherence directly, i.e. the maximal value ($\ell_\infty$) of $\abs{A^TA}-\abs{G_{ETF}}$ as appears in Eq.~\eqref{eq:regularization}, should be the preferred option. 

\begin{table}
\caption{Comparison of ETF optimization criteria on LeNet5 FC layer and Fashion MNIST} \label{table:FashionMnistLosses}
\begin{center}
 \begin{tabular}{l l l l l} 
 \hline
 Regularization & None & \textbf{ETF max ($\ell_\infty$)} & ETF sum ($\ell_1$) & ETF squared ($\ell_2$) \\ [0.5ex] 
 \hline
 Test Accuracy & 88.36\% & \textbf{90.89}\% & 88.89\% & 89.22\%\\
 \hline
\end{tabular}
\end{center}
\end{table}

An intuition behind the usage of the $\ell_\infty$ norm in the optimization is related to the concept of hard example mining. The loss focuses on the two columns that cause the Gram matrix to be the farthest from the one of an ETF.
It is known that in non-convex optimization, one may achieve improvement when focusing on the optimization of the harder examples, which improves the convergence and results \cite{triplet_loss,focal_loss}.

\section{Experiments} \label{Experiments}
We turn to evaluate our method apart of and on top of Dropout in the classification regime. Our analysis above applies to the regression (auto-encoder) case, and therefore it does not necessarily imply that in the classification case Dropout will encourage an ETF structure. Thus, we check here whether adding such a regularization can help also for classification. 
We emphasize that we do not try to compete with Dropout, nor we try to reach state-of-the-art results. In addition, the training time using the coherence is much higher than using Dropout. We simply aim at demonstrating the impact of ETF regularization with and without Dropout. The value of $\beta$ is chosen by cross-validation. To isolate the effect of the two methods, no other regularization techniques are used. We demonstrate our proposed strategy on FC layers, convolutional layers, and LSTM. Four known datasets (Fashion MNIST, CIFAR-10, Tiny-ImageNet, Penn tree bank) are used with their appropriate architectures.

Fashion MNIST \cite{FashionMnist} is a dataset similar to MNIST but with fashion related classes that are harder to classify compared to the standard MNIST. We use for it a LeNet5 based model.

CIFAR-10 is composed of 10 classes of $32\times32$ RGB natural images with 50,000 training images, and 10,000 testing images. The architecture used is also based on a variant of Lenet5.

Tiny Imagenet is composed of 200 classes of natural images with 500 training examples per class, and 10,000 images for validation. Each image is an RGB image of size $64{\times}64$. It is tested by top-1 and top-5 accuracy.
The architecture we use is an adaptation of the VGG-16 model \cite{VGG16} to the Tiny Imagenet dataset \cite{VGG16_detail}.

We perform word level prediction experiments on the Penn Tree Bank data set \cite{b9}. It consists of 929,000 training words, 73,000 validation words, and 82,000 test words. The vocabulary has 10,000 words. In this dataset, we measure the results by the attained perplexity, which we aim at reducing.
The architecture used is as in \cite{b8}. Two models are considered, where all of them involve LSTMs with two-layer, which are unrolled for $35$ steps. The $small$ model includes 200 hidden units, and the $medium$ includes 650.

The full implementation details appear in the sup. mat. and the code is available at: https://github.com/dorbank/An-ETF-view-of-Dropout-Regularization.

\begin{table}
\caption{FC layer regularization effect on test accuracy} \label{table:FC_layer}
\begin{center}
\resizebox{\columnwidth}{!}{%
 \begin{tabular}{P{2.5cm} P{2.5cm} P{2.5cm} P{2.5cm} P{2.5cm}}
 \hline
 Regularization & Fashion MNIST & CIFAR-10 & Tiny ImageNet (top1) & Tiny ImageNet (top5) \\ [0.5ex] 
 \hline
 None & 88.36\% & 84.41\% & 39.92\% & 65.29\%\\
 Dropout & 90.16\% & 86.16\% & 48.35\% & 73.13\%\\ 
  ETF & 90.89\% & 86.14\% & 44.21\% & 69.34\%\\
 \textbf{Dropout+ETF} & \textbf{91.91\%} & \textbf{86.94\%} & \textbf{49.78\%} & \textbf{73.55\%}\\
 \hline
\end{tabular}
}
\end{center}
\end{table}

\noindent {\bf Fully connected layers.}
We start by applying our ETF regularization on the FC layers on three image classification datasets: Fashion MNIST , CIFAR-10 and Tiny ImageNet.
As can be seen in Table~\ref{table:FC_layer}, the ETF regularization  improves the test accuracy, with and without Dropout. Note that it always improves the results of Dropout when combined together with it and that on the Fashion MNIST data it gets better performance also when it is used alone.

\noindent {\bf Convolutional layers.} Next, we apply our ETF regularization on the convolutional layers (Table \ref{table:conv_layer}). 
It can be observed that the ETF regularization has less effect on the convolutional layers compared to the FC ones, both when applied with and without Dropout. We conjecture that for classification tasks, the kernels of the different channels have already lower coherence than the columns of a FC weight matrix. It might be also that a regularization of the coherence of the stride matrix may lead to better results. 

\begin{table}
\caption{Convolution layer regularization effect on test accuracy} \label{table:conv_layer}
\begin{center}
\resizebox{\columnwidth}{!}{%
 \begin{tabular}{P{2.5cm} P{2.5cm} P{2.5cm} P{2.5cm} P{2.5cm}}
 \hline
 Regularization & Fashion MNIST & CIFAR-10 & Tiny ImageNet (top1) & Tiny ImageNet (top5) \\ [0.5ex] 
 \hline
 None & 88.36\% & 84.41\% & 39.92\% & 65.29\%\\
 Dropout & 91.14\% & 85.75\% & 43.44\% & 69.03\%\\ 
 ETF & 90.30\% & 85.15\% & 42.13\% & 67.05\%\\
 \textbf{Dropout+ETF} & \textbf{91.58\%} & \textbf{86.36\%} & \textbf{45.55\%} & \textbf{69.80\%}\\
 \hline
\end{tabular}
}
\end{center}
\end{table}

\begin{table}
\caption{LSTM layer regularization effect on test accuracy -  Penn Tree Bank dataset} \label{table:lstm_layer}
\begin{center}
\resizebox{\columnwidth}{!}{%
 \begin{tabular}{P{2.5cm} P{2.5cm} P{2.5cm} P{2.5cm} P{2.5cm}}
 \hline
 Regularization & small model (Val Perp.) & small model (Test Perp.) & medium model (Val Perp.) & medium model (Test Perp.) \\ [0.5ex] 
 \hline
 None & 121.39 & 115.91 & 123.012 & 122.853\\
 Dropout & 98.260 & 93.927 & 87.059 & 83.059\\
 ETF  & 104.425 & 99.398 & 115.868 & 111.956 \\
 \textbf{Dropout + ETF} & \textbf{93.998} & \textbf{90.139} & \textbf{85.267} & \textbf{81.646}\\
 \hline
\end{tabular}
}
\end{center}
\end{table}

\noindent {\bf LSTM.}
Lastly, we apply our ETF regularization on LSTM cells. We test it for both the small sized model and the medium sized one (Table \ref{table:lstm_layer}).
Notice that in this case, we also see the positive effect of the ETF regularization mainly when combined with Dropout. When applied alone, its effect is weaker in the medium model compared to the small one though it always leads to improvement. We believe that this difference should be further investigated.

\section{Conclusions}

This work provides a novel interpretation of the role of Dropout by bringing together two, similar but "unacquainted", research fields, namely, deep learning and frame theory. This combination provides the understanding that Dropout promotes an ETF structure when applied on a linear encoder in an autoencoder model. We have shown that adding a regularization that encourages an ETF structure improves the performance in these networks. The fact that in semi-supervised learning, the encoder also serves many times as a feature extractor for classification tasks, has led us to the usage of this ETF regularization also in standard neural networks, e.g., for classification, along with Dropout. This combination has shown improvement in different tasks and network types.
It showed that the bias induced by the proposed regularization is related to the one of Dropout. We believe that the relationship that this work draws between the two (both theoretically and empirically) should be further explored.

It appears that the study of frames can help to gain a better understanding of the Dropout regularization. We believe that this paper makes the first steps in this direction by studying the optimal frame created by Dropout in an autoencoder architecture that has a linear encoder.  
The improvement demonstrated in this work by the ETF regularization together with Dropout, for various tasks such as classification, suggests that the role of ETF in neural network optimization should be more deeply analyzed in these contexts.

\section{Acknowledgments}
We thank Prof. Ram Zamir and Marina Haikin for fruitful discussion and for introducing us to the analog coding setup. This work is supported by the ERC-STG SPADE grant.
\\
\\
\\
\\

\appendix{\huge \textcolor{black}{\textbf{Appendix}}}

\section{Dropout regularization}

$Dropout$ regularization, introduced by Srivastava et al. \cite{Dropout}, is one of the most popular regularization strategies. When applied on a given layer, it randomly drops hidden nodes along with their connections \cite{b5}. During training, in each batch, neurons are multiplied by a Bernoulli($p$) variable, which causes them to nullify with a probability $q = 1 - p$. Each weight connected to a nullified neuron does not influence the output, thus, it is not updated by backpropagation at the given training step. Clearly, the remaining weights are trained regularly. At test time, all outputs are multiplied by $p$ to sustain the overall weight norm.

Besides its great success in improving the generalization of neural networks, Dropout is also beneficial to avoid saddle points during training due to it stochasticity. Moreover, it incurs only a small additional complexity since it is linear in the features dimension.

Jindal et al. have used Dropout to deal with noisy labels, by reducing the certainty of a trained model and making it more robust. After the usual softmax layer, which is located at the end of the network, they have added a fully connected layer with Dropout, followed by another softmax layer. This led to a smoother label distribution for each sample, which is  less affected by the noisy labels \cite{noisyLabels}.

Note that each Dropout operation leads to a different approximation of the output of the network and therefore to a different optimization step. Thus, it is suggested that LSTM and GRU cells should have the same units dropped at each time step so the prediction would be consistent with respect to time  \cite{Dropout_RNN}.

{\bf Implicit bias.} Mianjy et al. \cite{b2}  study the implicit bias of Dropout \cite{b2}. It focuses on the case of a shallow network with a single hidden layer. Denote its matrices as $A \in \mathbb{R}^{m_1 \times n}$ and $B \in \mathbb{R}^{m_2 \times n}$. By applying Dropout with probability $p$ and using the squared loss, their optimization objective reads as:
\begin{equation}
\label{eq:dropout_objective_linear}
    f(A,B)=\mathbb{E}_{b_t \sim Ber(p),x \sim \D} \left[ \norm{y-\frac{1}{p}B\diag(b)A^Tx}^2 \right],
\end{equation}
where $\D$ is the distribution of $x$. Notice that when $p=1$, we simply get the case without dropout denoted as:
\begin{equation}
    \ell(A,B)= \mathbb{E}_{,x \sim \D} \left[ \norm{y-BA^Tx}^2 \right].
\end{equation}
First they show that the Dropout objective in \eqref{eq:dropout_objective_linear} can be rewritten as:
\begin{equation} \label{eq:dropout}
    f(A,B)=\ell(A,B) + \lambda \sum_{i=1}^n\norm{a_i}\norm{b_i},
\end{equation}
where $\lambda=\frac{1-p}{p}$, and $a_i$ and $b_i$ represent the $i^{th}$ columns of $A$ and $B$ respectively.
They note that this is equivalent to the square of the convex Path-Regularization \cite{path_regularization}, which is the square-root summation over all paths in the network, where in each path the squared weights product is calculated. In addition, they argue that the additional term of \eqref{eq:dropout} is an explicit instantiation of the implicit bias of dropout. They then define the term of jointly equalized matrices, where matrices $A$ and $B$ are considered jointly equalized iff $\forall i, \norm{a_i}\norm{b_i}=\norm{a_1}\norm{b_1}$. This notation is used where it is proven that if $(A,B)$ is a global minimum of \eqref{eq:dropout}, then $A$ and $B$ are jointly equalized. A clear but important observation, is that when $y=x$ and $m_1=m_2$ (the dimensions of $A$ and $B$ are equal) we get an objective of an autoencoder.

{\bf DeCov.} Cogswell et al. \cite{DeCov} use the fact that Dropout leads to less correlated features. Denote $h^n \in \mathbb{R}^d$ as the activations at a given hidden layer, where $n \in {1,...,N}$ refers to an index of one example from a batch of size $N$. The covariances between all pairs of activations $i$ and $j$ form the matrix $C$:
\begin{equation}
    C_{i,j}=\frac{1}{N}\sum_n{(h^n_i-\mu_i)(h^n_j-\mu_j)},
\end{equation}
where $\mu_i=\frac{1}{N}\sum_n{h^n_i}$ is the sample mean of activation $i$ over the batch.
The matrix $C$ is used in the DeCov \cite{DeCov} regularization, which explicitly regularizes the covariance of the features with respect to the training data by adding the following loss:
\begin{equation}
    \LL_{DeCov}=\frac{1}{2}(\norm{C}^2_F-\norm{\diag(C)}^2_2), 
\end{equation}
where $\norm{\cdot}$ is the frobenius norm, and $\diag(\cdot)$ returns the diagonal of a matrix.
Though useful, the connection between it and Dropout is not fully established. Moreover, DeCov is even shown to be adversarial to Dropout on some occasions.

\section{Autoencoders}

Autoencoders have been first introduced in \cite{AutoEncoder_original} as a neural network that is trained to reconstruct its input. Their main purpose is learning in an unsupervised manner an ``informative'' representation of the data that can be used for clustering. The problem, as formally defined in \cite{AutoEncoders_explanation}, is to learn the functions $A: \mathbb{R}^n \rightarrow \mathbb{R}^p$ (encoder) and $B: \mathbb{R}^p \rightarrow \mathbb{R}^n$ (decoder) that satisfy
\begin{equation}
\argmin_{A,B}\mathbb{E}[\Delta(x, B\circ A (x)],
\end{equation}
where $\Delta$ is an arbitrary distortion function, which is set to be the $\ell_2$-norm in our case, and $\mathbb{E}$ is the expectation over the distribution of $x$.


In the most popular form of autoencoders, $A$ and $B$ are neural networks \cite{NNAutoEncoder}.
In the special case that $A$ and $B$ are linear operations, we get a linear autoencoder \cite{linear_AutoEncoders}.

Since in training one may just get the identity operator for $A$ and $B$, which keeps the achieved representation the same as the input, some additional regularization is required. One option is to make the dimension of the representation smaller than the input. Another option is using  denoising autoencoders \cite{Denoising_AutoEncoders}. In these architectures, the  input is disrupted by some noise (e.g., additive white Gaussian noise or erasures using Dropout) and the autoencoder is expected to reconstruct the clean version of the input.


Another major improvement in the representation capabilities of autoencoders has been achieved by the variational autoencoders \cite{VariationalAutoEncoder}. These encode the input to latent variables, which represent the distribution that the data came from, and decode it by learning the posterior probability of the output from it.


Autoencoders may be trained in an end-to-end manner or gradually layer by layer. In the latter case,   they are ''stacked'' together, which leads to a deeper encoder. In \cite{ConvAutoEncoder}, this is done with convolutional autoencoders, and in \cite{Stacked_autoEncoders} with denoising ones.

\subsection{Use of autoencoders for classification.}

Autoencoders are also used for classification by using the encoder as a feature extractor and "plugging" it into a classification network. This is mainly done in the semi-supervised learning setup. First, the autoencoders are trained as described above in an unsupervised way. Then (or in parallel), the encoder is used as the first part of a classification network, and its weights may be fine tuned or not vary during training \cite{VariationalAutoEncoder}.
Notice that different types of autoencoders may be mixed to form new ones, as in  \cite{VariationalAutoEncoderClassification}, which uses them for classification, captioning, and unsupervised learning.

\section{The connection between ETF and MANOVA}
\begin{figure}
    \centering
    \includegraphics[scale=0.5]{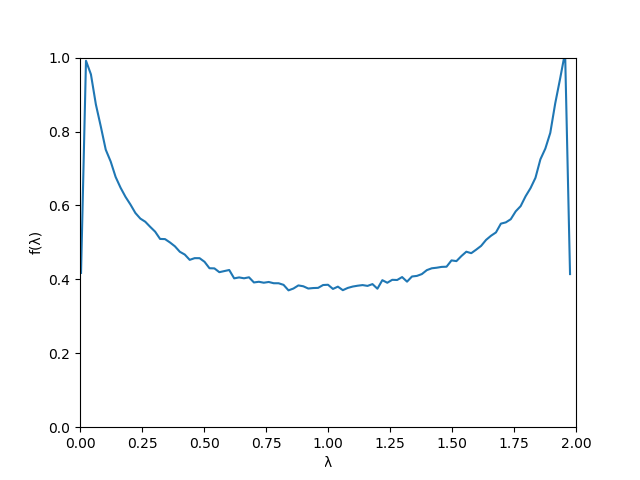}
    \caption{Empirical eigenvalue distribution of submatrices sampled from an ETF with $\beta=0.8$ and $\gamma=0.5$.}
    \label{fig:ETF_dist}
\end{figure}

It has been demonstrated in \cite{Marina_ETFManova} that frames that reach the Welch bound (also known as Equiangular Tight Frames(ETF)), have MANOVA distribution. The eigenvalue distribution of the submatrices of an ETF is shown empirically to resemble the MANOVA distribution (see Fig.~\ref{fig:ETF_dist} as an example).

In a following work \cite{Marina_FrameMoments}, the relationship between the MANOVA distribution and ETFs has been further supported by showing similarity between the moments of the MANOVA distribution and the ones of the ETF. The $d$-th moment of a random subset in $F$ is defined as
\begin{equation}\label{eq:moments}
m_d \overset{\Delta}{=} \frac{1}{n}\mathbb{E}[Tr((FPF^T)^d)],
\end{equation}
where $Tr(\cdot)$ is the trace operator and $P$ is a diagonal matrix with independent Bernoulli($p$) elements on its diagonal. 

It has been proven for $d=2,3,4$ that these moments are lower bounded by the moments of matrices with the Wachter's classical MANOVA distribution, plus a vanishing term (as n goes to infinity with $\frac{m}{n}$ held constant). The bound is proven to hold with equality for ETFs, where in the case of $d=4$ it is shown that it holds only for ETFs.
This leads us to assume that the subsets of ETF matrices indeed have MANOVA distribution.

\section{Regularizing convolution kernels}

\begin{figure}
    \centering
    \includegraphics[scale = 0.4]{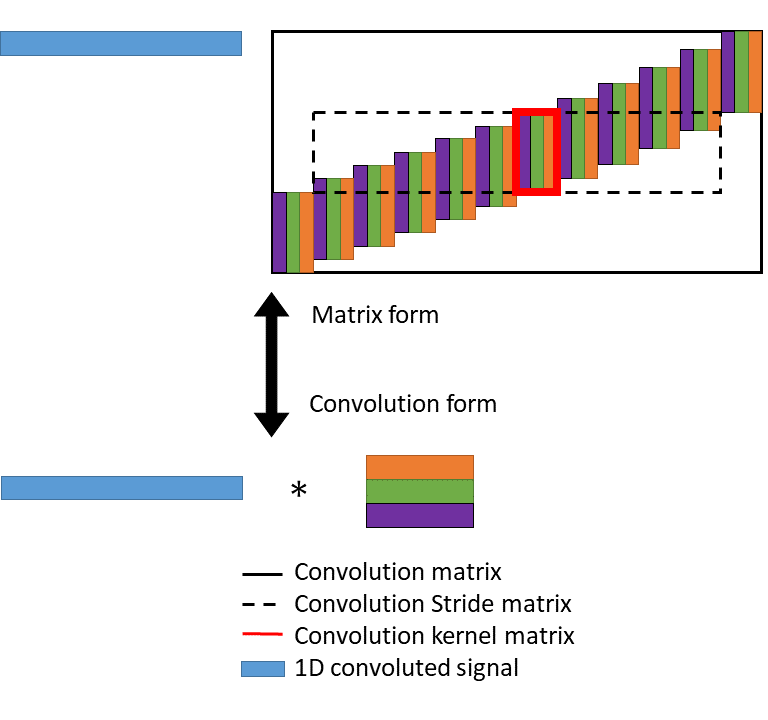}
    \caption{An illustration (1D case) of the equivalence between a convolution with three kernels and a multiplication with the equivalent Toeplitz matrix. Notice that the coherence of the convolution Toeplitz matrix is the same as the coherence of the smaller convolution stride matrix (marked by dashed lines).}
    \label{fig:Toeplitz}
\end{figure}

To apply our regularization on convolutional layers, we may use their corresponding convolution Toeplitz matrix as illustrated in Fig.~\ref{fig:Toeplitz}. Notice that the coherence of the convolution Toeplitz matrix is the same as the coherence of the smaller convolution stride matrix (marked by dashed lines) and thus, we apply the regularization directly on the stride convolution matrix. Yet, for simplicity, we just regularize the coherence between the convolution kernels (the matrix marked in red in Fig.~\ref{fig:Toeplitz}), which is the central part of the stride matrix. 
In the multi-dimensioanl case, each kernel is column-stacked and treated as a column vector in the regularized matrix.

\section{Implementation details}

\paragraph{Fashion MNIST.} The Fashion MNIST \cite{FashionMnist} is a dataset similar to MNIST but with fashion related classes that are harder to classify compared to the standard MNIST. It is composed of 60,000 examples as the train set, and 10,000 as the test set. Each example is a $28{\times}28$ grayscale image, associated with a label from 10 fashion related classes.

The architecture we used is based on LeNet5, and was changed a bit to examine a case where $m>n$. The FC layers were changed from $400\rightarrow120\rightarrow84\rightarrow10$ to $400\rightarrow800\rightarrow10$.
For the FC layers, the ETF parameter was set to 100, and the Dropout to 0.5. For the convolutional layers, when used as a sole regularizer, the ETF parameter was increased to 1000. The batch size was 128 and the score was taken as the best one in 400 epochs. The optimizer used was ADAM with a learning rate that diminished from $10^{-3}$ to $10^{-5}$.

\paragraph{CIFAR-10.} The CIFAR-10 dataset is composed of 10 classes of natural images with 50,000 training images, and 10,000 testing images. Each image is an RGB image of size $32{\times}32$.

The architecture is based on a variant of Lenet5 for this data set. It involves $5{\times}5{\times}32$ and $5{\times}5{\times}64$ convolution layers with $2{\times}2$ max pooling, followed by two FC layers of $1600\rightarrow1024\rightarrow10$. For the FC layers, the ETF parameter was set to 10 when it is the sole regularizer, and to 1 when combined with Dropout. For the convolutional layers, it was set to 10. The Dropout parameter was 0.5. The batch size was 128 and the score was taken as the best one in 300 epochs. The optimizer used was Nesterov Momentum with a momentum parameter of 0.9 and a learning rate that diminished from $10^{-2}$ to $10^{-3}$.

\paragraph{Tiny ImageNet.} The Tiny Imagenet dataset is composed of 200 classes of natural images with 500 training examples per class, and 10,000 images for validation. Each image is an RGB image of size $64{\times}64$. It is tested by top-1 and top-5 accuracy.

The architecture we use is an adaptation of the VGG-16 model \cite{VGG16} to the Tiny Imagenet dataset \cite{VGG16_detail}. It consists of ten $3{\times}3$ convolution layers, separated to four parts: two layers with 64 feature maps, two with 128, three with 256, and three with 512. All parts are separated by a $2{\times}2$ max pool, and after the last convolutional layer there is no pooling but FC layers of $25088\rightarrow4096\rightarrow2048\rightarrow200$. 

For the FC layers, the ETF parameter was set to 10 when it is the sole regularizer, and to 1 when combined with Dropout. For convolutional layers, it was set to 1. The Dropout parameter was 0.5. The batch size was 64 and the score was taken as the best one in 50 epochs. The optimizer used was Nesterov Momentum with a momentum parameter of 0.9 and a learning rate that diminished from $10^{-2}$ by a factor of $5$ when the validation top-1 accuracy ceased increasing.

\paragraph{Convolution regularization details.} In the convolution regularization experiments, we apply the dropout and ETF regularization as follows. For Fashion MNIST and CIFAR 10, we apply the regularization  on the second convolutional layer - right before the FC ones. 
For Tiny ImageNet, we apply it on the last three convolution layers - the ones with feature maps of size $512$. Dropout in all cases is applied once after the convolutional layers. In the case of the first two networks, it is applied after the pooling operation that follows the convolutions (since this gives better performance with Dropout).

\paragraph{Penn Tree Bank.} We perform word level prediction experiments on the Penn Tree Bank data set \cite{b9}. It consists of 929,000 training words, 73,000 validation words, and 82,000 test words. The vocabulary has 10,000 words. In this data set, we measure the results by the attained perplexity, which we aim at reducing. 

The architecture is as described in \cite{b8}. Two models are considered, where all of them involve LSTMs with two-layer, which are unrolled for $35$ steps. The $small$ model includes 200 hidden units, and the $medium$ includes 650. 

Small model parameters: When used as a sole regularizer, the ETF parameter was set to $1$, and when combined with Dropout, to $0.1$. The Dropout was set to 0.75. The score was taken as the best one in $30$ epochs on the validation set. The optimizer used was stochastic gradient descent (SGD) and the learning rate diminished from 1 by a factor of 0.7.

Medium model parameters: When used as a sole regularizer, the ETF parameter was set to $50$, and when combined with Dropout, to $1$. The Dropout was set to 0.5. The score was taken as the best one in 45 epochs on the validation set. The optimizer used was SGD and the learning rate diminished from 1 by a factor of 0.8.

\bibliography{ref}

\begin{thebibliography}{74}
\providecommand{\natexlab}[1]{#1}
\providecommand{\url}[1]{\texttt{#1}}
\expandafter\ifx\csname urlstyle\endcsname\relax
  \providecommand{\doi}[1]{doi: #1}\else
  \providecommand{\doi}{doi: \begingroup \urlstyle{rm}\Url}\fi

\bibitem[VGG(2017)]{VGG16_detail}
{VGG} code for tiny imagenet, 2017.
\newblock https://github.com/pat-coady/tiny\_imagenet.

\bibitem[Achille and Soatto(2018)]{denoising_AE}
A.~Achille and S.~Soatto.
\newblock Information dropout: Learning optimal representations through noisy
  computation.
\newblock \emph{IEEE Transactions on Pattern Analysis and Machine
  Intelligence}, 2018.

\bibitem[Baldi and Hornik(1989)]{linear_AutoEncoders}
P.~Baldi and K.~Hornik.
\newblock Neural networks and principal component analysis: Learning from
  examples without local minima.
\newblock \emph{Neural Netw.}, 2\penalty0 (1):\penalty0 53--58, January 1989.
\newblock ISSN 0893-6080.
\newblock \doi{10.1016/0893-6080(89)90014-2}.
\newblock URL \url{http://dx.doi.org/10.1016/0893-6080(89)90014-2}.

\bibitem[Baldi(2012)]{AutoEncoders_explanation}
Pierre Baldi.
\newblock Autoencoders, unsupervised learning, and deep architectures.
\newblock In Isabelle Guyon, Gideon Dror, Vincent Lemaire, Graham Taylor, and
  Daniel Silver, editors, \emph{Proceedings of ICML Workshop on Unsupervised
  and Transfer Learning}, volume~27 of \emph{Proceedings of Machine Learning
  Research}, pages 37--49, Bellevue, Washington, USA, 02 Jul 2012. PMLR.

\bibitem[Baldi and Sadowski(2013)]{b4}
Pierre Baldi and Peter~J Sadowski.
\newblock Understanding dropout.
\newblock In C.~J.~C. Burges, L.~Bottou, M.~Welling, Z.~Ghahramani, and K.~Q.
  Weinberger, editors, \emph{Advances in Neural Information Processing Systems
  26}, pages 2814--2822. Curran Associates, Inc., 2013.

\bibitem[Bodmann and Paulsen(2005)]{erase1}
B.~G. Bodmann and V.~I. Paulsen.
\newblock Frames, graphs and erasures.
\newblock \emph{Linear Algebra and its Applications}, 404:\penalty0 118 -- 146,
  2005.

\bibitem[Bodmann et~al.(2008)Bodmann, Casazza, and Balan]{frames1}
Bernhard~G. Bodmann, Pete Casazza, and Radu Balan.
\newblock Frames for linear reconstruction without phase.
\newblock \emph{The 42nd Annual Conference on Information Sciences and
  Systems}, pages 721--726, 2008.

\bibitem[Casazza and Kutyniok(2008)]{erase3}
Pete~G. Casazza and Gitta Kutyniok.
\newblock Robustness of fusion frames under erasures of subspaces and of local
  frame vectors.
\newblock \emph{Contemporary Mathematics}, 25:\penalty0 114--132, 2008.

\bibitem[Casazza and Kova{\v{c}}evi{\'{c}}(2003)]{UTF}
Peter~G. Casazza and Jelena Kova{\v{c}}evi{\'{c}}.
\newblock Equal-norm tight frames with erasures.
\newblock \emph{Advances in Computational Mathematics}, 18\penalty0
  (2):\penalty0 387--430, Feb 2003.

\bibitem[Casazza et~al.(2008)Casazza, Kutyniok, and Li]{frames2}
Peter~G Casazza, Gitta Kutyniok, and Shidong Li.
\newblock Fusion frames and distributed processing.
\newblock \emph{Applied and Computational Harmonic Analysis}, 25:\penalty0
  114--132, 2008.

\bibitem[Cavazza et~al.(2018)Cavazza, Morerio, Haeffele, Lane, Murino, and
  Vidal]{b11}
Jacopo Cavazza, Pietro Morerio, Benjamin Haeffele, Connor Lane, Vittorio
  Murino, and Rene Vidal.
\newblock Dropout as a low-rank regularizer for matrix factorization.
\newblock \emph{Proceedings of the Twenty-First International Conference on
  Artificial Intelligence and Statistics}, 84:\penalty0 435--444, 09--11 Apr
  2018.

\bibitem[Christensen(2003)]{frames3}
Ole Christensen.
\newblock An introduction to frames and riesz bases.
\newblock \emph{Birkhauser}, 2003.

\bibitem[Cogswell et~al.(2015)Cogswell, Ahmed, Girshick, Zitnick, and
  Batra]{DeCov}
Michael Cogswell, Faruk Ahmed, Ross~B. Girshick, Larry Zitnick, and Dhruv
  Batra.
\newblock Reducing overfitting in deep networks by decorrelating
  representations.
\newblock \emph{CoRR}, 2015.

\bibitem[Cotfas and Gazeau(2010)]{TF}
Nicolae Cotfas and Jean~Pierre Gazeau.
\newblock Finite tight frames and some applications.
\newblock \emph{Journal of Physics A: Mathematical and Theoretical},
  43\penalty0 (19):\penalty0 193001, 2010.
\newblock URL \url{http://stacks.iop.org/1751-8121/43/i=19/a=193001}.

\bibitem[Donoho and Elad(2003)]{Donoho02Optimally}
D.L. Donoho and M.~Elad.
\newblock Optimally sparse representation in general (nonorthogonal)
  dictionaries via $\ell^1$ minimization.
\newblock \emph{Proc. Nat. Aca. Sci.}, 100\penalty0 (5):\penalty0 2197--2202,
  March 2003.

\bibitem[Duarte-Carvajalino and Sapiro(2009)]{Carvajalino09Learning}
J.~M. Duarte-Carvajalino and G.~Sapiro.
\newblock Learning to sense sparse signals: Simultaneous sensing matrix and
  sparsifying dictionary optimization.
\newblock \emph{IEEE Transactions on Image Processing}, 18\penalty0
  (7):\penalty0 1395--1408, July 2009.

\bibitem[Elad(2007)]{Elad07Optimized}
M.~Elad.
\newblock Optimized projections for compressed sensing.
\newblock \emph{IEEE Transactions on Signal Processing}, 55\penalty0
  (12):\penalty0 5695--5702, Dec 2007.

\bibitem[Eldar(2015)]{Eldar15Sampling}
Y.~C. Eldar.
\newblock \emph{Sampling Theory: Beyond Bandlimited Systems}.
\newblock Cambridge University Press, 2015.

\bibitem[{Erd{\H o}s} and {Farrell}(2013)]{MANOVA}
L.~{Erd{\H o}s} and B.~{Farrell}.
\newblock {Local Eigenvalue Density for General MANOVA Matrices}.
\newblock \emph{Journal of Statistical Physics}, 152:\penalty0 1003--1032,
  September 2013.
\newblock \doi{10.1007/s10955-013-0807-8}.

\bibitem[Frazier{-}Logue and Hanson(2018)]{delta_rule}
Noah Frazier{-}Logue and Stephen~Jos{\'{e}} Hanson.
\newblock Dropout is a special case of the stochastic delta rule: faster and
  more accurate deep learning.
\newblock \emph{CoRR}, abs/1808.03578, 2018.

\bibitem[Gal and Ghahramani(2016{\natexlab{a}})]{Dropout_Bayesian_approx}
Yarin Gal and Zoubin Ghahramani.
\newblock Dropout as a bayesian approximation: Representing model uncertainty
  in deep learning.
\newblock In \emph{Proceedings of the 33rd International Conference on
  International Conference on Machine Learning - Volume 48}, ICML'16, pages
  1050--1059, 2016{\natexlab{a}}.

\bibitem[Gal and Ghahramani(2016{\natexlab{b}})]{Dropout_RNN}
Yarin Gal and Zoubin Ghahramani.
\newblock A theoretically grounded application of dropout in recurrent neural
  networks.
\newblock In \emph{NIPS}, 2016{\natexlab{b}}.

\bibitem[Ge et~al.(2018)Ge, Huang, Dong, and Scott]{triplet_loss}
Weifeng Ge, Weilin Huang, Dengke Dong, and Matthew~R. Scott.
\newblock Deep metric learning with hierarchical triplet loss.
\newblock In Vittorio Ferrari, Martial Hebert, Cristian Sminchisescu, and Yair
  Weiss, editors, \emph{Computer Vision -- ECCV 2018}, pages 272--288, Cham,
  2018. Springer International Publishing.

\bibitem[Ghiasi et~al.(2018)Ghiasi, Lin, and Le]{DropBlock}
Golnaz Ghiasi, Tsung-Yi Lin, and Quoc~V. Le.
\newblock Dropblock: A regularization method for convolutional networks.
\newblock In \emph{Proceedings of the 32nd International Conference on Neural
  Information Processing Systems}, NIPS’18, page 10750–10760, Red Hook, NY,
  USA, 2018. Curran Associates Inc.

\bibitem[Goodfellow et~al.(2016)Goodfellow, Bengio, and
  Courville]{Goodfellow16DeepLearningBook}
Ian Goodfellow, Yoshua Bengio, and Aaron Courville.
\newblock \emph{Deep Learning}.
\newblock MIT Press, 2016.

\bibitem[Haikin and Zamir(2016)]{Marina_AnalogCoding}
M.~Haikin and R.~Zamir.
\newblock Analog coding of a source with erasures.
\newblock In \emph{2016 IEEE International Symposium on Information Theory
  (ISIT)}, pages 2074--2078, July 2016.

\bibitem[Haikin et~al.(2017)Haikin, Zamir, and Gavish]{Marina_ETFManova}
Marina Haikin, Ram Zamir, and Matan Gavish.
\newblock Random subsets of structured deterministic frames have {MANOVA}
  spectra.
\newblock \emph{Proceedings of the National Academy of Sciences}, 114\penalty0
  (26):\penalty0 E5024--E5033, 2017.

\bibitem[Haikin et~al.(2018)Haikin, Zamir, and Gavish]{Marina_FrameMoments}
Marina Haikin, Ram Zamir, and Matan Gavish.
\newblock Frame moments and welch bound with erasures.
\newblock \emph{CoRR}, abs/1801.04548, 2018.

\bibitem[Han and Larson(2000)]{frames4}
Deguang Han and David~R. Larson.
\newblock Frames, bases and group representations.
\newblock \emph{American Mathematical Society,}, 697, 2000.

\bibitem[Hara et~al.(2017)Hara, Saitoh, and Shouno]{Dropout_ensemble}
Kazuyuki Hara, Daisuke Saitoh, and Hayaru Shouno.
\newblock Analysis of dropout learning regarded as ensemble learning.
\newblock \emph{CoRR}, abs/1706.06859, 2017.

\bibitem[Helmbold and Long(2015)]{b12}
David~P. Helmbold and Philip~M. Long.
\newblock On the inductive bias of dropout.
\newblock \emph{Journal of Machine Learning Research}, 16:\penalty0 3403--3454,
  2015.

\bibitem[Hinton et~al.(2012{\natexlab{a}})Hinton, Srivatava, Krizhvsky,
  Sutskever, and Salakhutdinov]{b5}
G.~E. Hinton, N.~Srivatava, A.~Krizhvsky, I.~Sutskever, and R.~R.
  Salakhutdinov.
\newblock Improving neural networks by preventing co-adaptation of feature
  detectors.
\newblock \emph{NIPS}, 2012{\natexlab{a}}.

\bibitem[Hinton et~al.(2012{\natexlab{b}})Hinton, Deng, Yu, Dahl, Mohamed,
  Jaitly, Senior, Vanhoucke, Nguyen, Sainath, and
  Kingbury]{Hinton2012DeepRecognition}
Geoffrey Hinton, Li~Deng, Dong Yu, George Dahl, Abdel-Rahman Mohamed, Navdeep
  Jaitly, Andrew Senior, Vincent Vanhoucke, Patrick Nguyen, Tara Sainath, and
  Brian Kingbury.
\newblock Deep neural networks for acoustic modeling in speech recognition.
\newblock \emph{IEEE Signal Processing Magazine}, 29\penalty0 (6):\penalty0
  82--97, 2012{\natexlab{b}}.

\bibitem[Ioffe and Szegedy(2015)]{BatchNorm}
Sergey Ioffe and Christian Szegedy.
\newblock Batch normalization: Accelerating deep network training by reducing
  internal covariate shift.
\newblock In \emph{Proceedings of the 32Nd International Conference on
  International Conference on Machine Learning - Volume 37}, ICML'15, pages
  448--456. JMLR.org, 2015.

\bibitem[Jindal et~al.(2016)Jindal, Nokleby, and Chen]{noisyLabels}
Ishan Jindal, Matthew~S. Nokleby, and Xuewen Chen.
\newblock Learning deep networks from noisy labels with dropout regularization.
\newblock \emph{2016 IEEE 16th International Conference on Data Mining (ICDM)},
  pages 967--972, 2016.

\bibitem[Kim(2014)]{Kim2014ConvolutionalClassification}
Yoon Kim.
\newblock Convolutional neural networks for sentence classification.
\newblock In \emph{EMNLP}, pages 1746--1751, 2014.

\bibitem[Kingma and Welling(2013)]{VariationalAutoEncoder}
Diederik~P. Kingma and Max Welling.
\newblock Auto-encoding variational bayes.
\newblock \emph{CoRR}, abs/1312.6114, 2013.

\bibitem[Krizhevsky et~al.(2012)Krizhevsky, Sutskever, and
  Hinton]{Krizhevsky2012ImageNetNetworks}
Alex Krizhevsky, Ilya Sutskever, and Geoffrey~E Hinton.
\newblock Imagenet classification with deep convolutional neural networks.
\newblock In \emph{Advances In Neural Information Processing Systems}, pages
  1--9, 2012.

\bibitem[Krogh and Hertz(1992)]{WeightDecay}
Anders Krogh and John~A. Hertz.
\newblock A simple weight decay can improve generalization.
\newblock In \emph{Advances in Neural Information Processing Systems}, pages
  950--957, 1992.

\bibitem[Kukačka et~al.(2017)Kukačka, Golkov, and
  Cremers]{Kukacka17Regularization}
J.~Kukačka, V.~Golkov, and D.~Cremers.
\newblock Regularization for deep learning: A taxonomy.
\newblock In \emph{ArXiv preprint}, 2017.

\bibitem[Larson and Scholze(2015)]{Larson2015}
David Larson and Sam Scholze.
\newblock Signal reconstruction from frame and sampling erasures.
\newblock \emph{Journal of Fourier Analysis and Applications}, 21\penalty0
  (5):\penalty0 1146--1167, Oct 2015.

\bibitem[LeCun et~al.(1993)LeCun, Simard, and Pearlmutter]{EigenValue}
Yann LeCun, Patrice~Y. Simard, and Barak Pearlmutter.
\newblock Automatic learning rate maximization by on-line estimation of the
  hessian's eigenvectors.
\newblock In S.~J. Hanson, J.~D. Cowan, and C.~L. Giles, editors,
  \emph{Advances in Neural Information Processing Systems 5}, pages 156--163.
  Morgan-Kaufmann, 1993.

\bibitem[Lin et~al.(2017)Lin, Goyal, Girshick, He, and Doll{\'a}r]{focal_loss}
Tsung-Yi Lin, Priya Goyal, Ross~B. Girshick, Kaiming He, and Piotr Doll{\'a}r.
\newblock Focal loss for dense object detection.
\newblock \emph{2017 IEEE International Conference on Computer Vision (ICCV)},
  pages 2999--3007, 2017.

\bibitem[Marcus et~al.(1993)Marcus, Marcinkiewicz, and B.]{b9}
M.~P. Marcus, M.~A. Marcinkiewicz, and Santorini B.
\newblock Building a large annotated corpus of english: The penn treebank.
\newblock \emph{Computational Linguistics}, 19(2):\penalty0 313–--330, 1993.

\bibitem[Masci et~al.(2011)Masci, Meier, Cire{\c{s}}an, and
  Schmidhuber]{ConvAutoEncoder}
Jonathan Masci, Ueli Meier, Dan Cire{\c{s}}an, and J{\"u}rgen Schmidhuber.
\newblock Stacked convolutional auto-encoders for hierarchical feature
  extraction.
\newblock In Timo Honkela, W{\l}odzis{\l}aw Duch, Mark Girolami, and Samuel
  Kaski, editors, \emph{Artificial Neural Networks and Machine Learning --
  ICANN 2011}, pages 52--59, Berlin, Heidelberg, 2011. Springer Berlin
  Heidelberg.
\newblock ISBN 978-3-642-21735-7.

\bibitem[Mianjy et~al.(2018)Mianjy, Arora, and Vidal]{b2}
Poorya Mianjy, Raman Arora, and Rene Vidal.
\newblock On the implicit bias of dropout.
\newblock In \emph{ICML}, 2018.

\bibitem[Neyshabur et~al.(2015)Neyshabur, Tomioka, and
  Srebro]{path_regularization}
Behnam Neyshabur, Ryota Tomioka, and Nathan Srebro.
\newblock Norm-based capacity control in neural networks.
\newblock In Peter Grünwald, Elad Hazan, and Satyen Kale, editors,
  \emph{Proceedings of The 28th Conference on Learning Theory}, volume~40 of
  \emph{Proceedings of Machine Learning Research}, pages 1376--1401, Paris,
  France, 03--06 Jul 2015. PMLR.
\newblock URL \url{http://proceedings.mlr.press/v40/Neyshabur15.html}.

\bibitem[P.~Baldi(2014)]{b4a}
P.~Sadowski P.~Baldi.
\newblock The dropout learning algorithm.
\newblock \emph{Artifical Intelligence}, 210:\penalty0 78--122, 2014.

\bibitem[Pal et~al.(2020)Pal, Lane, Vidal, and Haeffele]{structured_dropout}
Ambar Pal, Connor Lane, Ren{\'{e}} Vidal, and Benjamin~D. Haeffele.
\newblock On the regularization properties of structured dropout.
\newblock \emph{CVPR}, 2020.

\bibitem[Papyan et~al.(2017)Papyan, Romano, and Elad]{Papyan17Convolutional}
V.~Papyan, Y.~Romano, and M.~Elad.
\newblock Convolutional neural networks analyzed via convolutional sparse
  coding.
\newblock \emph{Journal of Machine Learning Research (JMLR)}, \penalty0
  (18):\penalty0 1--52, 2017.

\bibitem[Papyan et~al.(2018)Papyan, Romano, Sulam, and
  Elad]{Papyan18Theoretical}
V.~Papyan, Y.~Romano, J.~Sulam, and M.~Elad.
\newblock Theoretical foundations of deep learning via sparse representations:
  A multilayer sparse model and its connection to convolutional neural
  networks.
\newblock \emph{IEEE Signal Processing Magazine}, 35\penalty0 (4):\penalty0
  72--89, July 2018.
\newblock ISSN 1053-5888.
\newblock \doi{10.1109/MSP.2018.2820224}.

\bibitem[Pu et~al.(2016)Pu, Gan, Henao, Yuan, Li, Stevens, and
  Carin]{VariationalAutoEncoderClassification}
Yunchen Pu, Zhe Gan, Ricardo Henao, Xin Yuan, Chunyuan Li, Andrew Stevens, and
  Lawrence Carin.
\newblock Variational autoencoder for deep learning of images, labels and
  captions.
\newblock In \emph{Advances in Neural Information Processing Systems 29: Annual
  Conference on Neural Information Processing Systems 2016, December 5-10,
  2016, Barcelona, Spain}, pages 2352--2360, 2016.

\bibitem[Ranzato et~al.(2007)Ranzato, Huang, Boureau, and LeCun]{NNAutoEncoder}
M.~Ranzato, F.~J. Huang, Y.~Boureau, and Y.~LeCun.
\newblock Unsupervised learning of invariant feature hierarchies with
  applications to object recognition.
\newblock In \emph{2007 IEEE Conference on Computer Vision and Pattern
  Recognition}, pages 1--8, June 2007.
\newblock \doi{10.1109/CVPR.2007.383157}.

\bibitem[Rifai et~al.(2011)Rifai, Vincent, Muller, Glorot, and
  Bengio]{Rifai11Contractive}
Salah Rifai, Pascal Vincent, Xavier Muller, Xavier Glorot, and Yoshua Bengio.
\newblock Contractive auto-encoders: Explicit invariance during feature
  extraction.
\newblock In \emph{International Conference on International Conference on
  Machine Learning (ICML)}, pages 833--840, 2011.

\bibitem[Rumelhart et~al.(1986)Rumelhart, Hinton, and
  Williams]{AutoEncoder_original}
D.~E. Rumelhart, G.~E. Hinton, and R.~J. Williams.
\newblock Parallel distributed processing: Explorations in the microstructure
  of cognition, vol. 1.
\newblock chapter Learning Internal Representations by Error Propagation, pages
  318--362. MIT Press, Cambridge, MA, USA, 1986.
\newblock ISBN 0-262-68053-X.
\newblock URL \url{http://dl.acm.org/citation.cfm?id=104279.104293}.

\bibitem[Rupf and Massey(2006)]{SubsetInterest}
M.~Rupf and J.~L. Massey.
\newblock Optimum sequence multisets for synchronous code-division
  multiple-access channels.
\newblock \emph{IEEE Trans. Inf. Theor.}, 40\penalty0 (4):\penalty0 1261--1266,
  September 2006.
\newblock ISSN 0018-9448.
\newblock \doi{10.1109/18.335940}.
\newblock URL \url{https://doi.org/10.1109/18.335940}.

\bibitem[Ryan and Debbah(2009)]{Vandermonde}
{\O}yvind Ryan and M{\'{e}}rouane Debbah.
\newblock Asymptotic behaviour of random vandermonde matrices with entries on
  the unit circle.
\newblock volume~55, pages 3115--3148, 2009.

\bibitem[Scardapane et~al.(2017)Scardapane, Comminiello, Hussain, and
  Uncini]{L1}
Simone Scardapane, Danilo Comminiello, Amir Hussain, and Aurelio Uncini.
\newblock Group sparse regularization for deep neural networks.
\newblock \emph{Neurocomputing}, 241:\penalty0 81 -- 89, 2017.
\newblock ISSN 0925-2312.

\bibitem[Simonian and Zisserman(2015)]{VGG16}
Karen Simonian and Andrew Zisserman.
\newblock Very deep convolutional networks for large-scale image recognition.
\newblock \emph{International Conference on Learning Representations (ICLR)},
  2015.

\bibitem[Sokolic et~al.(2017)Sokolic, Giryes, Sapiro, and
  Rodrigues]{Sokolic17Margin}
J.~Sokolic, R.~Giryes, G.~Sapiro, and M.~R.~D. Rodrigues.
\newblock Robust large margin deep neural networks.
\newblock \emph{IEEE Transactions on Signal Processing}, 65\penalty0
  (16):\penalty0 4265--4280, 2017.

\bibitem[Srivastava et~al.(2014)Srivastava, Hinton, Krizhevsky, Sutskever, and
  Salakhutdinov]{Dropout}
N.~Srivastava, G.~Hinton, A.~Krizhevsky, I.~Sutskever, and R.~Salakhutdinov.
\newblock Dropout: A simple way to prevent neural networks from overfitting.
\newblock \emph{Journal of Machine Learning Research}, 15, 2014.
\newblock 1929--1958.

\bibitem[Tang et~al.(2020)Tang, Wang, Xu, Shi, Xu, Xu, and Xu]{DisOut}
Yehui Tang, Yunhe Wang, Yixing Xu, Boxin Shi, Chao Xu, Chunjing Xu, and Chang
  Xu.
\newblock Beyond dropout: Feature map distortion to regularize deep neural
  networks.
\newblock In \emph{The Thirty-Fourth {AAAI} Conference on Artificial
  Intelligence, {AAAI} 2020, The Thirty-Second Innovative Applications of
  Artificial Intelligence Conference, {IAAI} 2020, The Tenth {AAAI} Symposium
  on Educational Advances in Artificial Intelligence, {EAAI} 2020, New York,
  NY, USA, February 7-12, 2020}, pages 5964--5971. {AAAI} Press, 2020.
\newblock URL \url{https://aaai.org/ojs/index.php/AAAI/article/view/6057}.

\bibitem[Vincent et~al.(2008)Vincent, Larochelle, Bengio, and
  Manzagol]{Denoising_AutoEncoders}
Pascal Vincent, Hugo Larochelle, Yoshua Bengio, and Pierre-Antoine Manzagol.
\newblock Extracting and composing robust features with denoising autoencoders.
\newblock In \emph{Proceedings of the 25th International Conference on Machine
  Learning}, ICML '08, pages 1096--1103, New York, NY, USA, 2008. ACM.
\newblock ISBN 978-1-60558-205-4.
\newblock \doi{10.1145/1390156.1390294}.
\newblock URL \url{http://doi.acm.org/10.1145/1390156.1390294}.

\bibitem[Vincent et~al.(2010)Vincent, Larochelle, Lajoie, Bengio, and
  Manzagol]{Stacked_autoEncoders}
Pascal Vincent, Hugo Larochelle, Isabelle Lajoie, Yoshua Bengio, and
  Pierre-Antoine Manzagol.
\newblock Stacked denoising autoencoders: Learning useful representations in a
  deep network with a local denoising criterion.
\newblock \emph{J. Mach. Learn. Res.}, 11:\penalty0 3371--3408, December 2010.
\newblock ISSN 1532-4435.
\newblock URL \url{http://dl.acm.org/citation.cfm?id=1756006.1953039}.

\bibitem[Voulodimos et~al.(2018)Voulodimos, Doulamis, Doulamis, and
  Protopapadakis]{Voulodimos2018DeepReview}
Athanasios Voulodimos, Nikolaos Doulamis, Anastasios Doulamis, and Eftychios
  Protopapadakis.
\newblock Deep learning for computer vision: A brief review.
\newblock \emph{Computational Intelligence and Neuroscience}, 2018.

\bibitem[Wager et~al.(2013)Wager, Wang, and Liang]{NIPS2013_4882}
Stefan Wager, Sida Wang, and Percy~S Liang.
\newblock Dropout training as adaptive regularization.
\newblock In \emph{Advances in Neural Information Processing Systems (NIPS)},
  pages 351--359, 2013.

\bibitem[Wager et~al.(2014)Wager, Fithian, Wang, and Liang]{NIPS2014_5502}
Stefan Wager, William Fithian, Sida Wang, and Percy~S Liang.
\newblock Altitude training: Strong bounds for single-layer dropout.
\newblock In \emph{Advances in Neural Information Processing Systems (NIPS)},
  pages 100--108, 2014.

\bibitem[Wan et~al.(2013)Wan, Zeiler, Zhang, LeCun, and Fergus]{DropConnect}
Li~Wan, Matthew Zeiler, Sixin Zhang, Yann LeCun, and Rob Fergus.
\newblock Regularization of neural networks using dropconnect.
\newblock In \emph{Proceedings of the 30th International Conference on
  International Conference on Machine Learning - Volume 28}, ICML’13, page
  III–1058–III–1066. JMLR.org, 2013.

\bibitem[Wang and Manning(2013)]{fast_dropout}
Sida Wang and Christopher Manning.
\newblock Fast dropout training.
\newblock In Sanjoy Dasgupta and David McAllester, editors, \emph{Proceedings
  of the 30th International Conference on Machine Learning}, volume~28 of
  \emph{Proceedings of Machine Learning Research}, pages 118--126, 2013.

\bibitem[Welch(1974)]{b6}
L.~Welch.
\newblock Lower bounds on the maximum cross correlation of signals.
\newblock \emph{IEEE Transactions on Information theory}, 20, 1974.

\bibitem[Xiao et~al.(2017)Xiao, Rasul, and Vollgraf]{FashionMnist}
Han Xiao, Kashif Rasul, and Roland Vollgraf.
\newblock Fashion-mnist: a novel image dataset for benchmarking machine
  learning algorithms.
\newblock \emph{CoRR}, abs/1708.07747, 2017.

\bibitem[Zaremba et~al.(2014)Zaremba, Sutskever, and Vinyals]{b8}
Wojciech Zaremba, Ilya Sutskever, and Oriol Vinyals.
\newblock Recurrent neural network regularization, 2014.
\newblock URL \url{https://arxiv.org/abs/1409.2329}.

\bibitem[Zhang et~al.(2016)Zhang, Pezeshki, Brakel, Zhang, Laurent, Bengio, and
  Courville]{Zhang16Towards}
Ying Zhang, Mohammad Pezeshki, Philemon Brakel, Saizheng Zhang, César Laurent,
  Yoshua Bengio, and Aaron~C Courville.
\newblock Towards end-to-end speech recognition with deep convolutional neural
  networks.
\newblock In \emph{Interspeech}, pages 410--414, 2016.

\bibitem[Zhang et~al.(2015)Zhang, Lee, and Jordan]{L1_improper}
Yuchen Zhang, Jason~D. Lee, and Michael~I. Jordan.
\newblock l1-regularized neural networks are improperly learnable in polynomial
  time.
\newblock \emph{CoRR}, abs/1510.03528, 2015.

\end{thebibliography}

\end{document}